\documentclass{article}

\pdfoutput=1

\usepackage{microtype}
\usepackage{graphicx}
\usepackage{subfigure}
\usepackage{enumitem}
\usepackage{booktabs}
\usepackage[margin=1in]{geometry}
\usepackage{authblk}

\usepackage{mathtools, amsthm, amsfonts, amssymb}
\usepackage{caption}
\usepackage{algorithm,algorithmic} 

\DeclareMathOperator*{\ex}{\mathbb{E}}
\DeclareMathOperator*{\argmin}{arg\,min}
\DeclareMathOperator*{\argmax}{arg\,max}
\newcommand\given[1][]{\:#1\vert\:}
\usepackage{hyperref}

\title{Active Ordinal Querying for Tuplewise Similarity Learning}

\allowdisplaybreaks

\begin{document}

\author[1]{Gregory Canal\footnote{Equal contribution.}}

\newcommand\CoAuthorMark{\footnotemark[\arabic{footnote}]}
\author[2]{Stefano Fenu\protect\CoAuthorMark}
\author[1]{Christopher Rozell}
\affil[1]{School of Electrical and Computer Engineering}
\affil[2]{School of Interactive Computing}
\affil[ ]{Georgia Institute of Technology, Atlanta, GA}
\affil[ ]{\{gregory.canal,sfenu3,crozell\}@gatech.edu}

\date{}
\maketitle
 
    

\begin{abstract}
Many machine learning tasks such as clustering, classification, and dataset search benefit from embedding data points in a space where distances reflect notions of relative similarity as perceived by humans. A common way to construct such an embedding is to request triplet similarity queries to an oracle, comparing two objects with respect to a reference. This work generalizes triplet queries to tuple queries of arbitrary size that ask an oracle to rank multiple objects against a reference, and introduces an efficient and robust adaptive selection method called InfoTuple that uses a novel approach to mutual information maximization. We show that the performance of InfoTuple at various tuple sizes exceeds that of the state-of-the-art adaptive triplet selection method on synthetic tests and new human response datasets, and empirically demonstrate the significant gains in efficiency and query consistency achieved by querying larger tuples instead of triplets.
\end{abstract}

\section{Introduction}

Similarity learning is the process of assigning point coordinates to objects in a dataset such that distances between objects in the learned space are consistent with notions of similarity as perceived by humans. While these objects usually exist in some high-dimensional space (e.g., images, audio), very often the semantic information humans attribute to these objects lies in a low-dimensional space (e.g., items, words). Once this low-dimensional embedding is learned, existing intelligent algorithms~\cite{jamieson2011active,canal2019active} can be used to search the dataset with query complexity scaling in the embedding dimension, allowing large datasets to be searched quickly in applications such as task selection for robot learning from demonstration \cite{argall2009survey}, object recognition \cite{Ferrari2004}, or image retrieval \cite{yang2010boosting}.

To construct such an embedding for a given set of objects, queries that capture the similarity statistics between the objects in question must be made to human experts. While there exist several types of similarity queries that can be made (e.g., relative attributes between objects \cite{parikh2011relative}), we  focus on relative similarity queries posed to an oracle comparing objects with respect to a ``head'' (i.e., reference) object. Relative similarity queries are useful because they gather object similarity information using only object identifiers rather than predetermined features or attributes, allowing similarity learning methods to be applied to any collection of uniquely identifiable objects.  In contrast, if a head object were not specified, an oracle would need to use a feature-based criterion for ranking the object set, which is not viable in many applications of interest (e.g., learning human preferences).
\begin{figure}[t]
\begin{center}
\subfigure[]{
\fbox{\includegraphics[width=.44\linewidth]{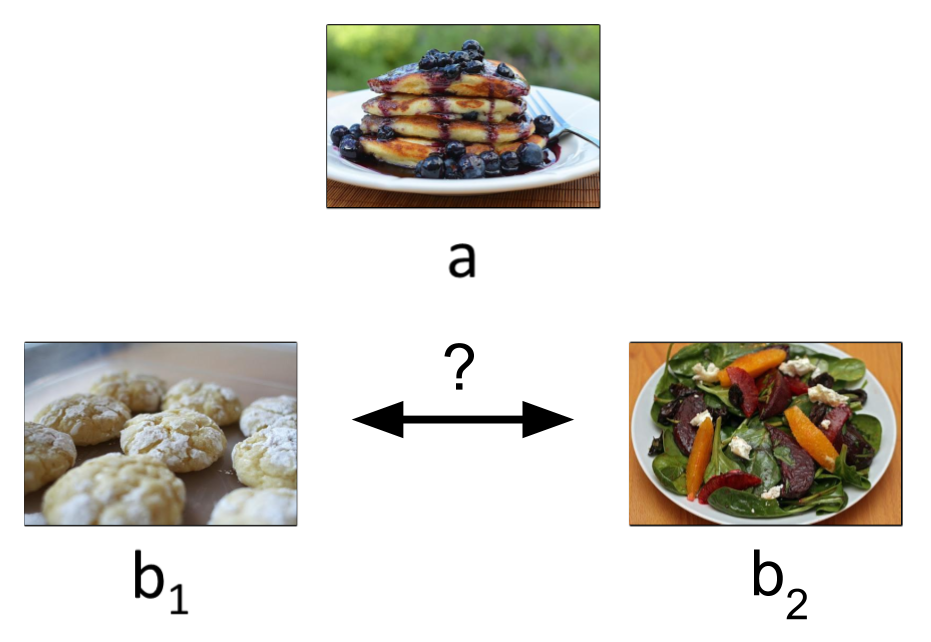}}
\label{fig:ambiguous_tuple}
}
\subfigure[]{
\fbox{\includegraphics[width=.44\linewidth]{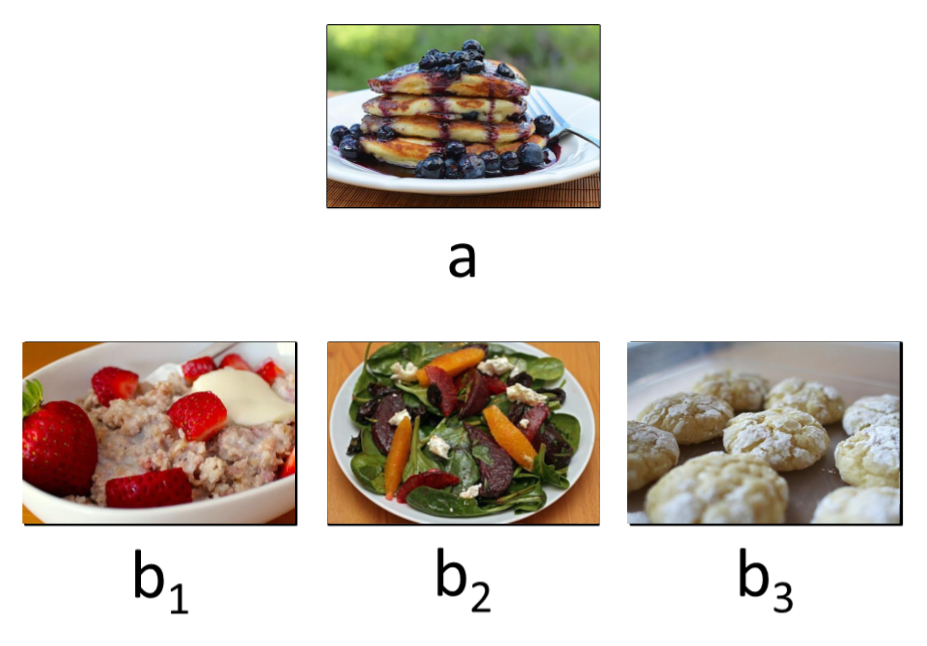}}
\label{fig:contextual_tuple}
}
\caption{In \ref{fig:ambiguous_tuple} it is ambiguous which item should be chosen as more similar to the head object, since both comparison items are similar in distinct ways. In \ref{fig:contextual_tuple}, adding one more comparison item can add context to disambiguate this choice.}
\label{fig:interface}
\end{center}
\end{figure}
Such relative similarity queries typically come in the form of triplet comparisons (i.e., ``is object $a$ more similar to object $b$ or $c$?'') \cite{tamuz2011adaptively,van2012stochastic,hoffer2015deep}. \emph{In our first main contribution, we extend these queries to larger rank orderings of tuples of $k$ objects to gather more information at once for similarity learning}. This query type takes the form ``rank objects $b_1$ through $b_{k-1}$ according to their similarity to object $a$.'' To the best of our knowledge, this study is the first attempt to leverage this generalized query type in similarity learning. The use of this query type is motivated by the fact that comparing multiple objects simultaneously provides increased context for a human expert \cite{fernando2015learning}, which can increase labeling consistency without a significant increase in human effort per query \cite{liang2014beyond} and has demonstrated benefits in settings such as rank learning \cite{cao2007learning}. In technical terms, tuplewise queries capture joint dependence between objects that isn't captured in triplet comparisons (which are often incorrectly modeled as independent queries). To illustrate this point, consider the difference between the triplet query and tuple query presented in Figure \ref{fig:interface}. In the triplet query,
multiple attributes could be used to rank a given query, increasing the ambiguity about which item should be chosen as more similar to the reference. Adding an item to the tuple can provide additional context about the entire dataset to the oracle, clarify which criterion should be used to rank the tuple and thereby making the query less ambiguous.

While tuple queries are appealing, their use presents two major challenges. First, in a dataset of $N$ objects queried with tuples of size $k$ there are $N\binom{N-1}{k-1}$ possible tuples. Labeling these individual tuples is prohibitively time consuming for large datasets. Even if uniformly random query selection is used to downsample this set, there is evidence that such a strategy is still punitively expensive \cite{jamieson2011low}. Requesting  an exhaustive number of queries is also inefficient from an information standpoint, since there is redundancy in the set of all tuple rankings. Second, in many settings of interest, the oracle answering such queries may be stochastic. For example, crowd oracles  may aggregate responses from experts with differing similarity judgements \cite{tamuz2011adaptively}, and individual oracles can be unreliable over time (especially for queries regarding similar objects). 

These issues can be ameliorated in part by leveraging tools from \textit{active learning}, the goal of which is to minimize the total labeling cost including the number of expert interactions (usually corresponding to monetary cost), aggregate response time, and computational cost needed to dynamically select queries. This is achieved through adaptive approaches that increase learning efficiency by using previous query responses to determine which information about a model is still ``missing'' as well as model the oracle's stochasticity. In this framework, unlabeled data points that optimize a measure of informativeness are selected for expert labeling.  One such metric, \textit{mutual information}, is a popular way to assess the reduction in uncertainty a query provides about unknown learning parameters \cite{settles2012active,lindley1956measure,mackay1992information}. In active similarity learning, the state-of-the-art is a strategy called ``Crowd Kernel Learning'' (CKL) that selects triplets that maximize the mutual information between a query response and the embedding coordinates of the head object \cite{tamuz2011adaptively}. However, CKL does not apply to ordinal queries of general tuples sizes ($k > 3$), and its formulation of mutual information only measures the information a query provides about the embedding coordinates of the head object, disregarding information about the locations of the other objects in the query. 

\emph{In our second main contribution, we address these deficiencies and the lack of an active similarity learning strategy for our new query type by introducing a novel method for efficient and robust adaptive selection of tuplewise queries of arbitrary size.} Our method, called \textit{InfoTuple}, maximizes the mutual information a query response provides about the \emph{entire} embedding, which is a direct measure of query informativeness that leverages the high degree of coupling between all of the objects in a query. InfoTuple relies on a novel set of simplifying yet reasonable assumptions for tractable mutual information estimation from a single batch of Monte Carlo samples. Our approach accounts for all objects in a query, while avoiding the need to decompose mutual information into a prohibitive number of terms. We demonstrate the performance of this method across datasets, oracle models, and tuple sizes, using both synthetic tests and newly collected large-scale human response datasets. In particular, we empirically show that InfoTuple's performance exceeds that of CKL and random queries, and furthermore that it benefits significantly from using larger tuples even after normalizing for tuple size. We also demonstrate the utility of our novel query type by showing an increase in query consistency for larger tuples over triplets, and show that these advantages can be gained without excessive labeling-time increases.

\section{Related Work}

Similarity learning from triplets is increasingly commonplace in modern AI, and popular deep learning architectures have been developed to leverage triplet labels \cite{hoffer2015deep}. Frameworks such as that of \cite{liu2012metric} or t-STE \cite{van2012stochastic} are relatively ubiquitous in the visualization community, and attempt to directly capture a notion of visual similarity close to that observed in psychometrics literature (e.g. \cite{chater1999scale}). However, for large datasets it is often punitively expensive to collect such exhaustive relationship data from labelers, so the development of approximate methods of learning such embeddings is a matter of interest to the AI community.

The bulk of the existing literature on active selection of ordinal queries for constructing these embeddings focuses on the case where distance relationships between objects can be determined with absolute certainty. This deterministic case is well studied, and lower bounds exist on the sample complexity needed to learn high-quality embeddings \cite{jamieson2011low}. In reality, responses are often not deterministic for a number of practical reasons  and probabilistic MDS methods have been proposed to model such cases \cite{tamuz2011adaptively}. Analytic results do exist characterizing bounds on prediction error in this setting \cite{jain2016finite}, but determining optimal strategies for query selection in the stochastic setting remains largely an open problem.

Specifically, to the best of our knowledge there have been no previous attempts to adaptively select relative comparisons with respect to a head object for general tuple sizes ($k\geq 3$) in the context of similarity learning. Prior work~\cite{liang2014beyond,yu2005svm} develops an active strategy for sampling tuples, but the query task is relative attribute ranking within the tuple according to some pre-specified attribute as opposed to comparison against a head object. Other work~\cite{qian2013active} actively samples the same query type as our study, but in the context of classification via label propagation. Research exists that is similar to our learning scenario since they actively sample tuples for relative similarity comparisons to a head for the sake of learning and searching an embedding of objects~\cite{cao2015facial}, but these comparisons are ternary `similar', `dissimilar', or `neither' labels and their methodology differs from the mutual information approach presented here. Similarly, other work~\cite{patterson2015tropel} actively samples tuplewise queries with binary `similar' or `dissimilar' label responses with respect to a head, but in the context of classification. Finally, the prior work~\cite{wilber2014cost} also employs such tuplewise binary queries for similarity learning, but with randomly selected queries. While no previous study addresses the similarity learning problem that we explore here, the existing literature demonstrates the effectiveness, efficiency, and feasibility of queries involving multiple objects and provides support for the practical use of our proposed query type.

\section{Methods}
\label{sec:methods}
The problem of adaptively selecting a tuplewise query can be formulated as follows: for a dataset $\mathcal{X}$ of $N$ objects, assume that there exists a $d$-dimensional vector of embedding coordinates for each object which are concatenated as columns in matrix $M \in \mathbb{R}^{d \times N}$. The \textit{similarity matrix} corresponding to $M$ is given by $K = M^T M$, which implies an $N \times N$ matrix $D$ of distances between the objects in $\mathcal{X}$. Specifically, the squared distance between the $i$th and $j$th objects in the dataset is given by $D^2_{i,j} = K_{i,i} - 2K_{i,j} + K_{j,j}$. These distances are assumed to be consistent in expectation with similarity comparisons from an \textit{oracle} (e.g., human expert or crowd) such that similar objects are closer and dissimilar objects are farther apart. Since relative similarity comparisons between tuples of objects inform their relative embedding distances rather than their absolute coordinates, our objective is to learn similarity matrix $K$ rather than $M$, which can be recovered from $K$ up to a change in basis \cite{tamuz2011adaptively}.

A tuplewise oracle query at time step $n$ is composed of a ``body'' of objects $B^n = \{b^n_1,b^n_2,\dots b^n_{k-1}\}$ which the oracle ranks by similarity with respect to some ``head'' object $a_n$. Letting $Q_n=\{a_n\}\cup B^n$ denote the $n$th posed tuple, we denote the oracle's ranking response as $R(Q_n)=\{R_1(Q_n),R_2(Q_n),\dots R_{k-1}(Q_n)\}$ which is a permutation of $B^n$ such that $R_1(Q_n) \prec R_2(Q_n) \dots \prec R_{k-1}(Q_n)$ where $b_i \prec b_j$ indicates that the oracle ranks object $b_i$ as more similar to $a_n$ than object $b_j$. Since the oracle is assumed to be stochastic, $R(Q_n)$ is a random permutation of $B^n$ governed by a distribution that is assumed to depend on $K$. This assumed dependence is natural because oracle consistency is likely coupled with notions of object similarity, and therefore with distances between the objects in $M$. The actual recorded oracle ranking is a random variate of $R(Q_n)$ denoted as $r(Q_n)$. Letting $r^n=\{r(Q_1),r(Q_2),\dots r(Q_n)\}$, define $\widehat{K}^n$ as an estimate of $K$ learned from previous rankings $r^n$, with corresponding distance matrix $\widehat{D}^n$.

Suppose that tuples $Q_1,Q_2,\dots Q_{n-1}$ have been posed as queries to the oracle with corresponding ranking responses $r^{n-1}$, and consider a Bayes optimal approach where after the $n$th query we estimate the similarity matrix as the maximum a-posteriori (MAP) estimator over a similarity matrix posterior distribution given by $f(K|r^n)$, i.e.\ $\widehat{K}_n = \argmax_K f(K|r^n)$ . To choose the query $Q_n$, a reasonable objective is to select a query that maximizes the achieved posterior value of the resulting MAP estimator (or equivalently one that maximizes the achieved logarithm of the posterior), corresponding to a higher level of confidence in the estimate. However, because the oracle response $r(Q_n)$ is unknown before a query is issued, the resulting maximized posterior value is unknown. Instead, a more reasonable objective is to select a query that maximizes the \textit{expected} value over the posterior of $R(Q_n)$. This can be stated as
\begin{equation*}
\argmax_{Q_n} \ex_{R(Q_n)}\left[\max_{K} \log f(K|R(Q_n),r^{n-1})\mid r^{n-1}\right].
\label{eq:argmaxQuery}
\end{equation*}
In practice, this optimization is infeasible since each expectation involves the calculation of several MAP estimates. Noting that maximization is lower bounded by expectation, this optimization can be relaxed by replacing the maximization over $K$ with an expectation over its posterior distribution given $R(Q_n)$ and $r^{n-1}$, resulting in a feasible maximization of a lower bound given by
\begin{equation}
\argmax_{Q_n} -h(K\mid R(Q_n),r^{n-1}) ,
\label{eq:negent}
\end{equation}
where $h(K\mid R(Q_n),r^{n-1})$ denotes \textit{conditional differential entropy} \cite{cover2012elements}.
Let the \textit{mutual information} between $K$ and $R(Q_n)$ given $r^{n-1}$ be defined by
\[
I(K;R(Q_n) \given r^{n-1}) = h(K\given r^{n-1}) - h(K \given R(Q_n),r^{n-1}),\]
and note that the second term is equal to (\ref{eq:negent}) while the first term does not depend on the choice of $Q_n$. Thus, maximizing (\ref{eq:negent}) over $Q_n$ is equivalent to maximizing $I(K,R(Q_n) \mid r^{n-1})$. Hence, we can adaptively select tuples that maximize mutual information as a means of greedily maximizing a lower bound on the log-posterior achieved by a MAP estimator, corresponding to a high estimator confidence.

However, calculating \eqref{eq:negent} for a candidate tuple is an expensive procedure that involves estimating the differential entropy of a combinatorially large number of posterior distributions, since the expectation with respect to $R(Q_n)$ is taken over $(k-1)!$ possible rankings. Instead, in the spirit of \cite{houlsby2012collaborative} we leverage the symmetry of mutual information to write the equivalent objective
\begin{equation}
\argmax_{Q_n} H(R(Q_n) \mid r^{n-1}) - H(R(Q_n) \mid K, r^{n-1})
\label{eq:objMI_alt}
\end{equation}
where $H(\cdot \mid \cdot)$ denotes conditional entropy of a discrete random variable. Estimating \eqref{eq:objMI_alt} for a candidate tuple only involves averaging ranking entropy over a \emph{single} posterior $f(K\mid r^{n-1})$, regardless of the value of $k$. This insight, along with suitable probability models discussed in the next sections, allows us to efficiently estimate mutual information for a candidate tuple over a single batch of Monte Carlo samples, rather than having to sample from $(k-1)!$ posteriors.

Furthermore, by interpreting entropy of discrete random variables as a measure of uncertainty, this form of mutual information maximization has a satisfying qualitative interpretation. The first entropy term in (\ref{eq:objMI_alt}) prefers tuples whose rankings are uncertain, preventing queries from being wasted on predictable or redundant responses. Meanwhile, the second term discourages tuples that have high expected uncertainty when conditioned on $K$; this prevents the selection of tuples that, even if $K$ were somehow revealed, would \textit{still} have uncertain rankings. Such queries are inherently ambiguous, and therefore uninformative to the embedding. Thus, maximizing mutual information optimizes the balance between these two measures of uncertainty and therefore prefers queries that are unknown to the learner but that can still be answered consistently by the oracle.

\subsection{Estimating Mutual Information}
\label{subsec:MI}
To tractably estimate the entropy terms in \eqref{eq:objMI_alt} for a candidate tuple, we employ several simplifying assumptions concerning the joint statistics of the query sequence and the embedding that allow for efficient Monte Carlo sampling:

\begin{enumerate}[label=(A\arabic*),align=left]
\setlength{\itemsep}{0pt}
    \item As is common in active learning settings, we assume that each query response $R(Q_n)$ is statistically independent of previous responses $r^{n-1}$, when conditioned on $K$.\label{assumption:condintrank}
    \item The distribution of $R(Q_n)$ conditioned on $K$ is only dependent on the distances between $a_n$ and the objects in $B^n$, notated as set $D_{Q_n}\coloneqq\{D_{a_n,b}:b\in B\}$. This direct dependence of tuple ranking probabilities on inter-object distances is rooted in the fact that the distance relationships in the embedding are assumed to capture oracle response behavior, and is a common assumption in ordinal embedding literature \cite{van2012stochastic,tamuz2011adaptively}. Furthermore, this conditional independence of $R(Q_n)$ from objects $x \not\in Q_n$ is prevalent in probabilistic ranking literature \cite{stern1990models}. In the next section, we describe a reasonable ranking probability model that satisfies this assumption. \label{assumption:R_distances}
	\item $D$ is conditionally independent of $r^{n-1}$, given $\widehat{D}^{n-1}$. This assumption is reasonable because embedding methods used to estimate $\widehat{K}^{n-1}$ (and subsequently $\widehat{D}^{n-1}$) are designed such that distances in the estimated embedding preserve the response history contained in $r^{n-1}$. In practice, it is more convenient to model an embedding posterior distribution by conditioning on $\widehat{D}^{n-1}$, learned from the previous responses $r^{n-1}$, rather than by conditioning on $r^{n-1}$ itself. This is in the same spirit of CKL, where the current embedding estimate is used to approximate a posterior distribution over points.\label{assumption:condintD}
    %
    \item Conditioned on $\widehat{D}^{n-1}$, the posterior distribution of $D_{Q_n}$ is normally distributed about the corresponding values in $\widehat{D}_{Q_n}^{n-1}$, i.e.\ $D^{n-1}_{a_n,b} \sim \mathcal{N}(\widehat{D}^{n-1}_{a_n, b},\sigma_{n-1}^2)~\forall b \in B$, where $\sigma_{n-1}^2$ is a variance parameter. Imposing Gaussian distributions on inter-object distances is a recent approach to modeling uncertainty in ordinal embeddings \cite{lohaus2019uncertainty} that allows us to approximate the distance posterior with a fixed batch of samples from a simple distribution. Furthermore, the combination of this model with \ref{assumption:R_distances} means that we only need to sample from the normal distributions corresponding to the objects in $Q_n$. We choose $\sigma_{n-1}^2$ to be the sample variance of all entries in $\widehat{D}^{n-1}$, which is a heuristic that introduces a source of variation that preserves the scale of the embedding. \label{assumption:normal}
\end{enumerate}
Combining these assumptions, with a slight abuse of notation by writing $H(X) = H(p(X))$ for a random variable $X$ with probability mass function $p(X)$, and $\mathcal{N}^{n-1}_{Q_n}$ to represent normal distribution $\mathcal{N}(\widehat{D}^{n-1}_{Q_n},\sigma_{n-1}^2)$, we have
\begin{align*}
    H(R(Q_n) \given r^{n-1}) &= H\left(\ex_K \left[p(R(Q_n)\given K,r^{n-1})\given r^{n-1}\right]\right)
    \\ &= H\left(\ex_K \left[p(R(Q_n)\mid K)\mid r^{n-1}\right]\right)&\text{\ref{assumption:condintrank}}
    \\ &= H\left(\ex_{D_{Q_n}} \left[p(R(Q_n)\mid D_{Q_n})\mid r^{n-1}\right]\right)&\text{\ref{assumption:R_distances}}
    \\ &= H\left(\ex_{D_{Q_n}} \left[p(R(Q_n)\mid D_{Q_n})\mid \widehat{D}^{n-1}\right]\right)&\text{\ref{assumption:condintD}}
    \\ &= H\left(\ex_{D_{Q_n}\sim \mathcal{N}^{n-1}_{Q_n}} \left[p(R(Q_n)\mid D_{Q_n})\right]\right)&\text{\ref{assumption:normal}}
\end{align*}
%
%
%
%
%
%
%
Similarly, we have
\[
H(R(Q_n) \given K, r^{n-1})=\ex_{D_{Q_n}\sim \mathcal{N}^{n-1}_{Q_n}} \left[H\left(p(R(Q_n)\given D_{Q_n})\right)\right]
\]
This formulation allows a fixed-sized batch of samples to be drawn and evaluated over, the size of which can be tuned based on real-time performance specifications. This enables us to separate our computational budget and mutual information estimation accuracy from the size of the tuple query.

\subsection{Embedding Technique}
\label{subsec:embed}
In order to maximize the flexibility of our approach and draw a closer one-to-one comparison to existing methods for similarity learning, we train our embedding on our actively selected tuples by first decomposing a tuple ranking into $k-2$ \emph{constituent triplets} defined by the set $\{R_i(Q_m) \prec R_{i+1}(Q_m) : 1 \leq i \leq k-2,~m\le n\}$, and then learning an embedding from these triplets with any triplet ordinal embedding algorithm of choice. Since we compare performance against CKL in our experiments, our proposed embedding technique follows directly from the probabilistic MDS formulation in \cite{tamuz2011adaptively} so as to evaluate the effectiveness of our novel query selection strategy in a controlled setting. We wish to constrain our learned similarity matrix to the set of symmetric unit-length PSD matrices, so we consider the set S of such matrices: $S=\{K\succeq 0 | K_{11}=K_{22}=\cdots=K_{NN}=1\}$. We denote the closest matrix in $S$ to $K$ as $P_S(K) = \argmin_{A\in S} \sum_{ij}(K_{ij} - A_{ij})^2$. Projecting to the element in $S$ closest to $K$ is a quadratic program, which we solve by gradient projection descent on $K$. We do this by selecting an initial $K^0$ arbitrarily, and for each iteration computing $K^{t+1} = P_S(K^t - \eta\nabla l_t(K^t))$ with $l_t$ being the empirical log-loss at iteration $t$ i.e.\ $l_t = \log\frac{1}{p}$, and $p$ being the probability that the oracle correctly ordered the constituent triplets of the selected tuples. For the response probability of an individual triplet, we adopt the model in \cite{tamuz2011adaptively} that is reminiscent of Bradley-Terry pairwise score models \cite{bradley1952rank}: for parameter $\mu>0$, $p(b_1 \prec b_2)=(D_{a,b_2}^2+\mu)/(D_{a,b_1}^2+D_{a,b_2}^2+2\mu)$.

\subsection{Tuple Response Model}

%
%
Our proposed technique is compatible with any tuple ranking model that satisfies \ref{assumption:R_distances}. However, since we use the triplet response model listed above in the probabilistic MDS formulation, combined with the need for a controlled test against CKL, we extend their model to the tuplewise case as follows: we first decompose an oracle's ranking into its constituent triplets, and then apply
\begin{equation*}
p(R(Q_n) \given D_{Q_n}) \propto \prod_{i=1}^{k-2}  \frac{D^2_{a, R_{i+1}(Q_n)} + \mu}{D^2_{a,R_{i}(Q_n)}+D^2_{a,R_{i+1}(Q_n)} + 2\mu },
\end{equation*}
for parameter $\mu > 0$. This model corresponds to oracle behavior that ranks objects proportionally to the ratio of their distances with respect to $a$, such that closer (resp.\ farther) objects are more (resp.\ less) likely to be deemed similar. Models of this type are generally held to be similar to the scale-invariant models present in some human perceptual systems \cite{chater1999scale}.

\subsection{Adaptive Algorithm}

Combining these concepts, we have the following algorithm titled InfoTuple, summarized in Algorithm \ref{algo:InfoTuple}: the algorithm requires that some initial set of randomly selected tuples be labeled to provide a reasonable initialization of the learned similarity matrix. Since the focus of this work is on the effectiveness of various adaptive selection methods, this initialization is standardized across methods considered in our results. Specifically, following established practice~\cite{tamuz2011adaptively}, a ``burn-in'' period is used where $T_0$ random triplets are posed for each object $a$ in object set $\mathcal{X}$, with $a$ being the head of each query. Then, for each time step $n$ we learn a similarity matrix $\widehat{K}^{n-1}$ on the set of previous responses $r^{n-1}$ by using probabilistic MDS. To make a comparison to CKL, we follow their procedure and subsequently pose a single tuple for each head $a \in \mathcal{X}$. However, it is possible to adaptively choose $a$ with our method by searching over both head and body objects for a maximally informative tuple. The body of each tuple, given some head $a$, is chosen by uniformly downsampling the set of possible bodies and selecting the one that maximizes the mutual information, calculated using the aforementioned probability model in our estimation procedure. This highlights the importance of computational tractability in estimating mutual information, since for a fixed computing budget per selected query, less expensive mutual information estimation allows for more candidate bodies to be considered. For a tuple size of $k$ we denote the run of an algorithm using that tuple size as InfoTuple-$k$.

\begin{algorithm}[t]
\caption{InfoTuple-$k$}
\label{algo:InfoTuple}
\begin{algorithmic}
	\REQUIRE object set $\mathcal{X}$, rate $\omega$, sample size $N_f$, horizon $T$
    \STATE $r^0 \leftarrow \emptyset$ initialize set of oracle responses
    \STATE $\widehat{K}^0 \leftarrow$ initialize embedding
    \FOR{$n = 1$ \TO $T$}
    	\STATE $\widehat{D}^{n-1} \leftarrow$ calculate pairwise distances from $\widehat{K}^{n-1}$
        \STATE $\sigma_{n-1}^2 \leftarrow \frac{1}{N^2}\sum_{d \in \widehat{D}^{n-1}} \Bigl(d - \frac{1}{N^2}\sum_{d \in \widehat{D}^{n-1}} d \Bigr)^2$
      \FORALL{$a \in \mathcal{X}$}
      	\STATE $\beta \leftarrow$ downsampled $k\!-\!1$ sized bodies at rate $\omega$
      	\FORALL{$B \in \beta$}
      	  \STATE $Q \leftarrow \{a\}\cup B$
      	  \STATE $D_s  \sim \mathcal{N}(\widehat{D}^{n-1}_Q,\sigma_{n-1}^2)$, drawn $N_f$ times
      	   \STATE 
      	   $I_B \leftarrow H\left(\sum\limits_{D \in D_s}\frac{p(R(Q)|D)}{N_f}\right) - \sum\limits_{D\in D_s}\frac{H(p(R(Q)|D))}{N_f}$
      	 \ENDFOR
      	\STATE $B \leftarrow \argmax_{B \in \beta}{I_B}$
        \STATE $r \leftarrow$ oracle ranks objects in $B$ relative to $a$
        \STATE $r^{n} \leftarrow r^{n-1} \cup r$
      \ENDFOR
      \STATE $\widehat{K}^n \leftarrow \operatorname{probabilisticMDS}(r^{n})$
    \ENDFOR
    \ENSURE $\widehat{K}^T$
\end{algorithmic}
\end{algorithm}

\section{Experiments}

Our results on synthetic and human response datasets show that InfoTuple's adaptive selection outperforms both random query selection and that of CKL\footnote{Code available at \url{https://github.com/siplab-gt/infotuple}}. This is true even when normalizing for changes in tuple size and when normalizing for labeling effort, showing that the incurred benefit is not only due to the increased information inherently present in larger tuples but also due to our improved adaptive selection. We also show that there are inherent consistency benefits to the use of larger queries, and that human labelers can respond to these query types in practice without undue cost.

\subsection{Datasets}
\label{sec:data}

To evaluate algorithm performance in a controlled setting, we constructed a synthetic evaluation dataset by generating a point cloud drawn from a $d$-dimensional multivariate normal distribution. To simulate oracle responses for this dataset, we use the popular Plackett-Luce permutation model to sample a ranking for a given head and body \cite{cao2007learning,guiver2009bayesian}. In this response model, each object in a tuple body is assigned a score according to a scoring function, which in our case is based on the distance in the underlying space between each object and the head. For a given subset of body objects, the probability of an object being ranked as most similar to the head is its score divided by the scores of all objects in that subset, and we generate each simulated oracle response by sequentially sampling objects without replacement from a tuple according to this model. We chose this tested response model to differ from the one we use to estimate mutual information in order to demonstrate the robustness of our method to mismatched noise models, and evaluate an additional Gaussian noise model in the appendix. This dataset was used to compare InfoTuple-3, InfoTuple-4, InfoTuple-5, CKL, Random-3, and Random-5 across noiseless, Gaussian, and Plackett-Luce oracles.
\begin{figure*}[t]
\centering
\subfigure[][]{
\includegraphics[height=1.59in]{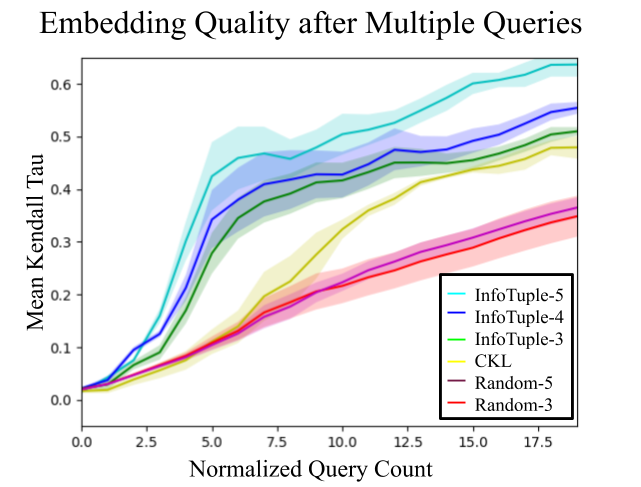}
\label{fig:baseline_comparison}
}
\subfigure[][]{
\includegraphics[height=1.59in]{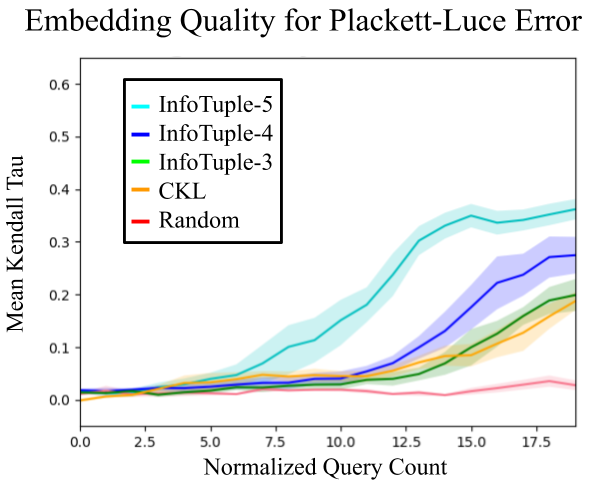}
\label{fig:power_law_error}
}
\subfigure[][]{
\includegraphics[height=1.59in]{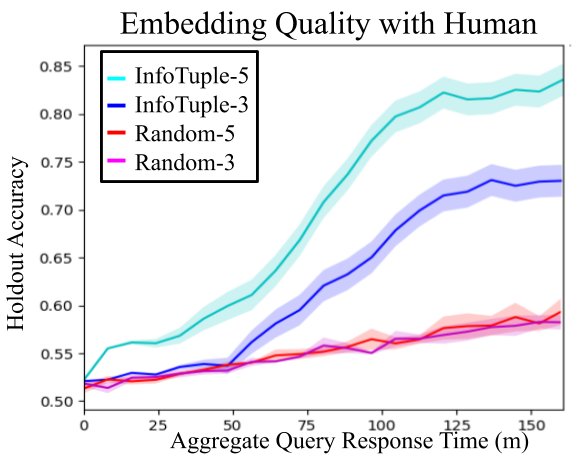}
\label{fig:empirical_results}
}
\caption{\ref{fig:baseline_comparison} and \ref{fig:power_law_error} show a comparison of the fidelity of the learned embedding to the ground truth embedding with a simulated deterministic (left) and a stochastic (right) oracle, plotted with $\pm 1$ standard error. Results shown are for a synthetic dataset of $N=500$ points from a two-dimensional dataset. \ref{fig:empirical_results} shows holdout accuracy on human-subject tests with $N=5000$.}
\end{figure*}

%
To demonstrate the broader applicability of our work in real-world settings and evaluate our proposed technique on perceptual similarity data, we also collected a large dataset of human responses to tuplewise queries through Amazon Mechanical Turk. Drawing 3000 food images from the Food-10k dataset \cite{wilber2015snack}, we presented over 7000 users with a total of 192,000 varying-size tuplewise queries chosen using Infotuple-3, InfoTuple-5, Random-3, and Random-5 as selection strategies across three repeated runs of each algorithm. Users were evaluated with one repeat query out of 25, and users who responded inconsistently to the repeat query were discarded. Query bodies were always shuffled when presented to minimize the impact of any possible order effect, and it was not found to be the case that there was any significant order effect in the human responses. Initial embeddings for each of these methods were trained on 5,000 triplet queries drawn from \cite{wilber2015snack}. Although experimental costs prevented us from extending the experiments in Figure \ref{fig:empirical_results} to larger tuple sizes, in order to verify the feasibility of having humans respond to larger tuples in practice we performed a separate data collection in which we asked users to rank randomly selected tuples up to a size of $k=10$ and recorded the labeling time for each response.

\subsection{Evaluation Metrics}

In order to directly measure the preservation of object rankings between the ground truth object coordinates and the embedding learned from oracle responses, we use Kendall's Tau rank correlation coefficient \cite{kendall1938new}. To get an aggregate measure of quality when comparing an estimated embedding to a ground-truth embedding, we take the mean of Kendall's Tau across the total rankings obtained by setting each object as the head and sorting all objects by embedding distance to the head. In our experiments with human respondents it is not possible to use this measure, as the ``ground truth'' embedding that corresponds to human preferences is not known. In these cases we instead measure the accuracy with respect to a held-out set of queries drawn from the Food-10k dataset \cite{wilber2015snack}, which is a common embedding quality metric \cite{van2012stochastic,wilber2015snack}. The holdout accuracy is the fraction of a held out set of triplet comparisons that agrees with distances in the final learned embedding. To capture a notion of the internal coherence between a set of oracle responses and an embedding that is learned from them, we measure the mean rank correlation between each response in this set and the ranking over the same objects imputed from the learned embedding--we refer to this as the \emph{coherence} of a set of tuples.

One issue that naturally arises when comparing results from strategies that select tuples of different size is normalization, as larger tuples will naturally be more informative. In human-response studies normalization is relatively straightforward, as we can simply normalize with respect to the total time spent labeling queries in order to reflect the total labeling cost. While other more comprehensive measures of labeler effort exist, labeling time is a first-order approximation for the cognitive load of a labeling task and is the most salient metric for determining the cost of a large-scale data collection. In the case of synthetic data, we instead compute a \emph{normalized query count} corresponding to the number of constituent triplet comparisons defining the relation of each body point to the head in the tuple. This is justified since in practice we decompose tuples in this way when feeding them into the embedding algorithm, and corresponds to the size of a tuple's transitive reduction (a common representation in learning-to-rank literature \cite{kingston2009comparing}). Additional experimental details such as hyperparameter selection are available in the appendix.

\subsection{Experimental Results}

Using simulated data, we show a direct comparison of embedding quality from using InfoTuple, CKL, and Random queries under a simulated deterministic oracle (Figure \ref{fig:baseline_comparison}) and two simulated stochastic oracles (Figure \ref{fig:power_law_error}), and note that InfoTuple consistently outperformed the other methods. We note two important observations from these results: first, regardless of the oracle used, larger tuple sizes for InfoTuple tended to perform better and converge faster than did smaller tuple sizes even after normalizing for the tuple size, showing the benefit of larger tuples beyond just providing more constituent triplets. Recalling that the Plackett-Luce oracle was not directly modeled in our estimate of mutual information, this lends support to the robustness of our technique to various oracle distributions. Second, results on Random-3, Random-4 and Random-5 are comparable, implying both that the improvements seen in InfoTuple are not solely due to the difference in tuple sizes and that our choice of normalization is appropriate. Note that since random query performance did not change with tuple size, Figure \ref{fig:power_law_error} only shows Random-3 for the sake of visual clarity.

\begin{figure}[t]
\centering
\includegraphics[width=0.4\linewidth]{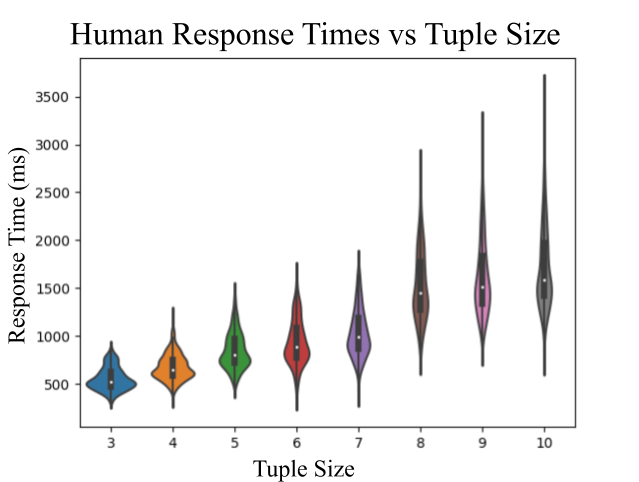}
\caption{This violin plot shows the distribution of timing responses for random queries from size $k=3$ to $k=10$, for the purpose of measuring labeler effort. The response time for $k\leq7$ shows only modest increases in cost, although responses above these sizes require significantly more effort.}
\label{fig:empirical_timing}
\end{figure}

Using the Mechanical Turk dataset described previously, we also show that these basic results extend to real data situations when the stochastic response model is not exactly known, and allows us to examine the complexity of acquiring data with increasing tuple sizes. While larger tuples sizes produce more informative queries, it is possible that the information gained incurs a hidden cost in the complexity or labeler effort involved in acquiring the larger query.  Specifically, it can be the case that maximizing query informativeness can produce queries that are more difficult to answer \cite{baldridge2009well}. Fortunately, the results on tuplewise comparisons collected for our Mechanical Turk dataset indicate that this is not an issue for our proposed use case. In particular, Figure \ref{fig:empirical_results} shows the accuracy results when predicting the labels from a held out set of 1200 triplet queries.  These results show an increase in the effectiveness of InfoTuple adaptive selection as well as increasing tuples sizes when plotted against the aggregate query response time. In other words, any increase in query complexity (measured by response time) is more than compensated for by the increased information acquired by the query and the increase in the resulting quality of the learned embedding.

Figure \ref{fig:empirical_timing} explores this issue further by examining the response times for our additional timing dataset as a function of query size. There are only modest increases in the ranking time cost with increasing tuple size, leading to the significant gains observed in normalized information efficiency in this range of tuple sizes. While it is true that complexity cost will continue to increase for larger tuple sizes and the gains in information efficiency are not guaranteed to increase indefinitely and there may also be additional factors in the choice of optimal tuple size for a given problem, we show that up to a modest tuple size it is strictly more useful to ask tuplewise queries than triplet queries.

\begin{figure}[t]
\centering
\includegraphics[width=0.4\linewidth]{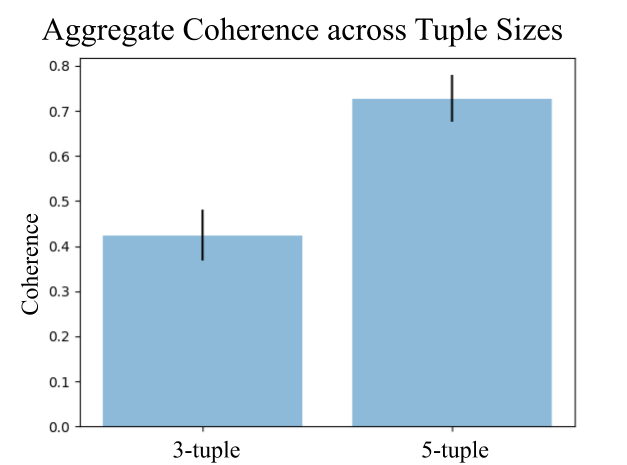}
\caption{Measuring the aggregate coherence for all tuples of size 3 and size 5 (i.e. over 80,000 tuples at each size) with respect to an aggregate embedding learned for each tuple size, we find that there is a significant difference in their internal coherence as measured by a t-test (p=0.007181). We hypothesize that the difference is due to an increase in context available to the oracle. Error bars depict $\pm 1$ standard error.}
\label{fig:coherence}
\end{figure}

One possible reason for why tuples outperform triplets is that asking a query that contains more objects provides additional context for the oracle about the contents of the dataset, allowing it to more reliably respond to ambiguous comparisons than if these were asked as triplet queries. As a result of this increase in context, oracles tend to respond to larger queries significantly more coherently than they do to smaller ones, as shown in Figure \ref{fig:coherence}. We note that this is not guaranteed to increase indefinitely as larger tuples are considered, but the effect is noticeable for modest increases in tuple sizes and is clear when comparing 5-tuples to triplets.
%

\section{Discussion}

In this paper we proposed InfoTuple, an adaptive tuple selection strategy based on maximizing mutual information for relative tuple queries for similarity learning. We introduce the tuple query for similarity learning, present a novel set of assumptions for efficient estimation of mutual information, and through the collection of new user-response datasets, provide new insights into the gains acquired by using larger tuples in learning efficiency and query consistency. After testing on synthetic and real datasets, InfoTuple was found to more effectively learn similarity-based object embeddings than random queries and state-of-the-art triplet queries for both synthetic data (with a typical oracle model) and in a real world experiment. The performance gains were especially evident for larger tuples and even after normalizing for tuple size, indicating that the proposed selection objective that maximizes the mutual information between the query response and the entire embedding yields information gains that are not simply due to an increase in tuple size. Taken together, these results suggest that large tuples selected with InfoTuple supply richer and more robust embedding information than their triplet and random counterparts.

In practice, larger tuple sizes can provide more context for the oracle, increasing the reliability of the responses without significant increases in labeling effort. In the pathological extreme, the level of effort almost certainly outweighs the benefits of larger tuples, as an oracle would have to provide a ranking over the entire dataset. Despite this downside in extreme tuple sizes, our human study results indicate that performance increases hold up in the real-world for moderate tuple sizes. This interesting tradeoff between informativeness per query and real-world oracle behavior merits a more comprehensive study on the psychometric aspects of the problem, in the spirit of \cite{miller1956magical}.

\section{Acknowledgements}
This work is partially supported by NSF CAREER award CCF-1350954, ONR grant number N00014-15-1-2619 and AFRL contract number FA8750-19-C-0200.

\appendix\clearpage
\section{Appendix}

\begin{figure}
\def \fh {2in}
\centering
\begin{minipage}[t]{0.45\textwidth}
\centering
\includegraphics[height=\fh]{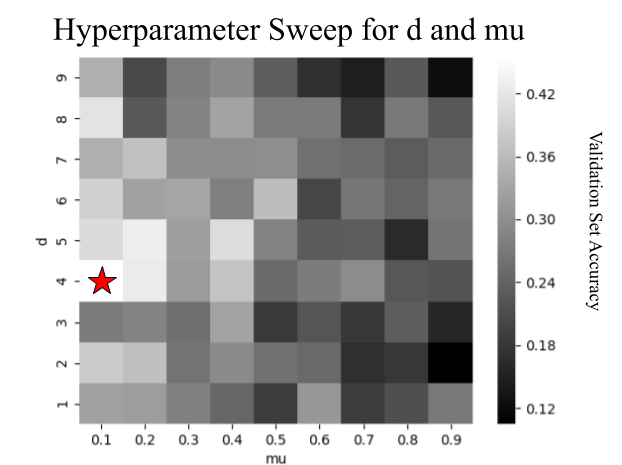}
\captionof{figure}{Hyperparameter sweep for Food10k dataset. Experimental values of $d=4$ and $\mu=0.1$ were found to be the most effective on a held-out validation set of triplets.}
\label{fig:hyperparameter}
\end{minipage}
\hfill
\begin{minipage}[t]{0.45\textwidth}
\centering
\includegraphics[height=\fh]{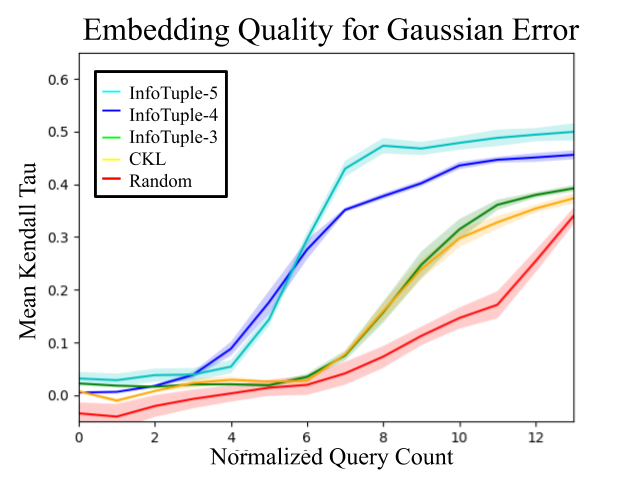}
\label{fig:gaussian_error.png}
\captionof{figure}{Synthetic experiment results using an oracle with Gaussian noise. Results were broadly consistent with those of the Plackett-Luce oracle in spite of the mismatch between the oracle noise and the embedding.}
\end{minipage}
\end{figure}

\subsection{Experimental Details}
 
 For each of the human-subject experiments, $\mu$ was set to 0.1 and $d$ was set to 4 per the hyperparameter search shown in Figure \ref{fig:hyperparameter}. The validation set for this search was an additional 500 heldout triplets from the Food10k dataset. In the synthetic experiments provided, $\mu$ was set to 0.5 and d was set to 2 to match the dimensionality of the generating distribution. The stochastic oracle had a high noise level, inverting 33\% of tuple responses. Higher tuple sizes were strongly correlated with both higher performance and higher robustness to error (even when normalized by the effective number of pairwise queries), indicating performance gains for InfoTuple that are not simply due to increasing tuple sizes. A heuristic was used to pick a number of samples for the Monte Carlo estimation of the mutual information, with $\frac{N}{10}$ samples being used in practice.
 
 Figure 2 in the paper body shows empirical performance for query selection algorithms on predicting labels from held out triplet queries in the Mechanical Turk dataset described. Experimental horizons for human subject experiments were chosen based on estimates of the initial steps of convergence and had to be limited due to high experimental costs. Turk subjects were presented with queries in batches of 25, with one repeated tuple across the batch as a test for validity. If the repeat query was not answered the same way by the user both times it was asked the batch was discarded. Order effects were controlled for by shuffling queries prior to presenting them to users for labeling, ensuring that any queries presented to multiple users would appear in different orders and that the test queries would also appear differently each time.
 
 \subsection{Oracle Details}
 
 Two different models of oracle noise were used in our synthetic experiments, Plackett-Luce noise and Gaussian noise. These models were chosen to be different from the one we use to estimate mutual information in order to demonstrate the robustness of our method. In the body of the paper we describe the selection process used by the Plackett-Luce oracle noise, which works by assigning latent scores to objects on the basis of their distances in some synthetic ``ground truth'' embedding space. The Gaussian noise model, instead of applying noise directly at the level of the ranking responses, applies noise at the level of the oracle's representation of the ``ground truth'' embedding by adding Gaussian noise to the coordinates of each point drawn from the ``ground truth'' embedding before imputing a ranking from distances in the oracle's noisy interpretation of the space. For the Plackett-Luce error model results shown in the paper body, 33\% of individual rankings were inverted.
 
\subsection{Computational Complexity}

The computational complexity of the embedding calculation is that of a typical MDS algorithm- for any $M \in \mathbb{R}^{d\times N}$ an approximate solution can be found in $O(N)$ for $d < N$ \cite{aflalo2013spectral}. Our case has an $N$ far greater than $d$ while still being of manageable size, allowing for a fast linear-time approximation.

With respect to the entropy calculation itself, the inner loop computing the mutual information from a given tuple is computable in $O(N_fk^2)$. However, the computational complexity for a given algorithm iteration is dominated by the $O(\omega{N \choose k-1})$ cost of generating and iterating over large pools of candidate tuples, meaning that the run-time is heavily dependent on the choice of the sampling rate $\omega$ and distance sample size $N_f$, and the question of how to efficiently estimate similar mutual information quantities without the use of Monte Carlo methods remains open.

\bibliographystyle{IEEEtran}
\bibliography{active_ordinal}

\begin{thebibliography}{10}
\providecommand{\url}[1]{#1}
\csname url@samestyle\endcsname
\providecommand{\newblock}{\relax}
\providecommand{\bibinfo}[2]{#2}
\providecommand{\BIBentrySTDinterwordspacing}{\spaceskip=0pt\relax}
\providecommand{\BIBentryALTinterwordstretchfactor}{4}
\providecommand{\BIBentryALTinterwordspacing}{\spaceskip=\fontdimen2\font plus
\BIBentryALTinterwordstretchfactor\fontdimen3\font minus
  \fontdimen4\font\relax}
\providecommand{\BIBforeignlanguage}[2]{{%
\expandafter\ifx\csname l@#1\endcsname\relax
\typeout{** WARNING: IEEEtran.bst: No hyphenation pattern has been}%
\typeout{** loaded for the language `#1'. Using the pattern for}%
\typeout{** the default language instead.}%
\else
\language=\csname l@#1\endcsname
\fi
#2}}
\providecommand{\BIBdecl}{\relax}
\BIBdecl

\bibitem{jamieson2011active}
K.~G. Jamieson and R.~Nowak, ``Active ranking using pairwise comparisons,'' in
  \emph{Advances in Neural Information Processing Systems}, 2011, pp.
  2240--2248.

\bibitem{canal2019active}
G.~Canal, A.~Massimino, M.~Davenport, and C.~Rozell, ``Active embedding search
  via noisy paired comparisons,'' in \emph{International Conference on Machine
  Learning}, 2019, pp. 902--911.

\bibitem{argall2009survey}
B.~D. Argall, S.~Chernova, M.~Veloso, and B.~Browning, ``A survey of robot
  learning from demonstration,'' \emph{Robotics and autonomous systems},
  vol.~57, no.~5, pp. 469--483, 2009.

\bibitem{Ferrari2004}
\BIBentryALTinterwordspacing
V.~Ferrari, T.~Tuytelaars, and L.~Van~Gool, \emph{Simultaneous object
  recognition and segmentation by image exploration}.\hskip 1em plus 0.5em
  minus 0.4em\relax Berlin, Heidelberg: Springer Berlin Heidelberg, 2004, pp.
  40--54. [Online]. Available:
  \url{http://dx.doi.org/10.1007/978-3-540-24670-1_4}
\BIBentrySTDinterwordspacing

\bibitem{yang2010boosting}
L.~Yang, R.~Jin, L.~Mummert, R.~Sukthankar, A.~Goode, B.~Zheng, S.~C. Hoi, and
  M.~Satyanarayanan, ``A boosting framework for visuality-preserving distance
  metric learning and its application to medical image retrieval,'' \emph{IEEE
  Transactions on Pattern Analysis and Machine Intelligence}, vol.~32, no.~1,
  pp. 30--44, 2010.

\bibitem{parikh2011relative}
D.~Parikh and K.~Grauman, ``Relative attributes,'' in \emph{Computer Vision
  (ICCV), 2011 IEEE International Conference on}.\hskip 1em plus 0.5em minus
  0.4em\relax IEEE, 2011, pp. 503--510.

\bibitem{tamuz2011adaptively}
O.~Tamuz, C.~Liu, S.~Belongie, O.~Shamir, and A.~T. Kalai, ``Adaptively
  learning the crowd kernel,'' \emph{arXiv preprint arXiv:1105.1033}, 2011.

\bibitem{van2012stochastic}
L.~Van Der~Maaten and K.~Weinberger, ``Stochastic triplet embedding,'' in
  \emph{Machine Learning for Signal Processing (MLSP), 2012 IEEE International
  Workshop on}.\hskip 1em plus 0.5em minus 0.4em\relax IEEE, 2012, pp. 1--6.

\bibitem{hoffer2015deep}
E.~Hoffer and N.~Ailon, ``Deep metric learning using triplet network,'' in
  \emph{International Workshop on Similarity-Based Pattern Recognition}.\hskip
  1em plus 0.5em minus 0.4em\relax Springer, 2015, pp. 84--92.

\bibitem{fernando2015learning}
B.~Fernando, S.~Gavves, D.~Muselet, and T.~Tuytelaars, ``Learning to rank based
  on subsequences,'' in \emph{Proceedings ICCV 2015}, 2015, pp. 2785--2793.

\bibitem{liang2014beyond}
L.~Liang and K.~Grauman, ``Beyond comparing image pairs: Setwise active
  learning for relative attributes,'' in \emph{Proceedings of the IEEE
  conference on Computer Vision and Pattern Recognition}, 2014, pp. 208--215.

\bibitem{cao2007learning}
Z.~Cao, T.~Qin, T.-Y. Liu, M.-F. Tsai, and H.~Li, ``Learning to rank: from
  pairwise approach to listwise approach,'' in \emph{Proceedings of the 24th
  international conference on Machine learning}.\hskip 1em plus 0.5em minus
  0.4em\relax ACM, 2007, pp. 129--136.

\bibitem{jamieson2011low}
K.~G. Jamieson and R.~D. Nowak, ``Low-dimensional embedding using adaptively
  selected ordinal data,'' in \emph{Communication, Control, and Computing
  (Allerton), 2011 49th Annual Allerton Conference on}.\hskip 1em plus 0.5em
  minus 0.4em\relax IEEE, 2011, pp. 1077--1084.

\bibitem{settles2012active}
B.~Settles, ``Active learning,'' \emph{Synthesis Lectures on Artificial
  Intelligence and Machine Learning}, vol.~6, no.~1, pp. 1--114, 2012.

\bibitem{lindley1956measure}
D.~V. Lindley, ``On a measure of the information provided by an experiment,''
  \emph{The Annals of Mathematical Statistics}, pp. 986--1005, 1956.

\bibitem{mackay1992information}
D.~J. MacKay, ``Information-based objective functions for active data
  selection,'' \emph{Neural computation}, vol.~4, no.~4, pp. 590--604, 1992.

\bibitem{liu2012metric}
E.~Y. Liu, Z.~Guo, X.~Zhang, V.~Jojic, and W.~Wang, ``Metric learning from
  relative comparisons by minimizing squared residual,'' in \emph{Data Mining
  (ICDM), 2012 IEEE 12th International Conference on}.\hskip 1em plus 0.5em
  minus 0.4em\relax IEEE, 2012, pp. 978--983.

\bibitem{chater1999scale}
N.~Chater and G.~D. Brown, ``Scale-invariance as a unifying psychological
  principle,'' \emph{Cognition}, vol.~69, no.~3, pp. B17--B24, 1999.

\bibitem{jain2016finite}
L.~Jain, K.~G. Jamieson, and R.~Nowak, ``Finite sample prediction and recovery
  bounds for ordinal embedding,'' in \emph{Advances In Neural Information
  Processing Systems}, 2016, pp. 2711--2719.

\bibitem{yu2005svm}
H.~Yu, ``Svm selective sampling for ranking with application to data
  retrieval,'' in \emph{Proceedings of the eleventh ACM SIGKDD international
  conference on Knowledge discovery in data mining}.\hskip 1em plus 0.5em minus
  0.4em\relax ACM, 2005, pp. 354--363.

\bibitem{qian2013active}
B.~Qian, X.~Wang, F.~Wang, H.~Li, J.~Ye, and I.~Davidson, ``Active learning
  from relative queries.'' in \emph{IJCAI}, 2013, pp. 1614--1620.

\bibitem{cao2015facial}
C.~Cao and H.-Z. Ai, ``Facial similarity learning with humans in the loop,''
  \emph{Journal of Computer Science and Technology}, vol.~30, no.~3, pp.
  499--510, 2015.

\bibitem{patterson2015tropel}
G.~Patterson, G.~Van~Horn, S.~J. Belongie, P.~Perona, and J.~Hays, ``Tropel:
  Crowdsourcing detectors with minimal training.'' in \emph{HCOMP}, 2015, pp.
  150--159.

\bibitem{wilber2014cost}
M.~J. Wilber, I.~S. Kwak, and S.~J. Belongie, ``Cost-effective hits for
  relative similarity comparisons,'' in \emph{Second AAAI Conference on Human
  Computation and Crowdsourcing}, 2014.

\bibitem{cover2012elements}
T.~M. Cover and J.~A. Thomas, \emph{Elements of information theory}.\hskip 1em
  plus 0.5em minus 0.4em\relax John Wiley \& Sons, 2012.

\bibitem{houlsby2012collaborative}
N.~Houlsby, F.~Huszar, Z.~Ghahramani, and J.~M. Hern{\'a}ndez-Lobato,
  ``Collaborative gaussian processes for preference learning,'' in
  \emph{Advances in neural information processing systems}, 2012, pp.
  2096--2104.

\bibitem{stern1990models}
H.~Stern, ``Models for distributions on permutations,'' \emph{Journal of the
  American Statistical Association}, vol.~85, no. 410, pp. 558--564, 1990.

\bibitem{lohaus2019uncertainty}
M.~Lohaus, P.~Hennig, and U.~von Luxburg, ``Uncertainty estimates for ordinal
  embeddings,'' \emph{arXiv preprint arXiv:1906.11655}, 2019.

\bibitem{bradley1952rank}
R.~A. Bradley and M.~E. Terry, ``Rank analysis of incomplete block designs: I.
  the method of paired comparisons,'' \emph{Biometrika}, vol.~39, no. 3/4, pp.
  324--345, 1952.

\bibitem{guiver2009bayesian}
J.~Guiver and E.~Snelson, ``Bayesian inference for plackett-luce ranking
  models,'' in \emph{proceedings of the 26th annual international conference on
  machine learning}.\hskip 1em plus 0.5em minus 0.4em\relax ACM, 2009, pp.
  377--384.

\bibitem{wilber2015snack}
\BIBentryALTinterwordspacing
M.~Wilber, I.~S. Kwak, D.~Kriegman, and S.~Belongie, ``Learning concept
  embeddings with combined human-machine expertise,'' in \emph{International
  Conference on Computer Vision (ICCV)}, 2015. [Online]. Available:
  \url{http://vision.cornell.edu/se3/wp-content/uploads/2015/09/main.pdf}
\BIBentrySTDinterwordspacing

\bibitem{kendall1938new}
M.~G. Kendall, ``A new measure of rank correlation,'' \emph{Biometrika},
  vol.~30, no. 1/2, pp. 81--93, 1938.

\bibitem{kingston2009comparing}
P.~Kingston and M.~Egerstedt, ``Comparing apples and oranges through partial
  orders: An empirical approach,'' in \emph{American Control Conference, 2009.
  ACC'09.}\hskip 1em plus 0.5em minus 0.4em\relax IEEE, 2009, pp. 5434--5439.

\bibitem{baldridge2009well}
J.~Baldridge and A.~Palmer, ``How well does active learning actually work?:
  Time-based evaluation of cost-reduction strategies for language
  documentation,'' in \emph{Proceedings of the 2009 Conference on Empirical
  Methods in Natural Language Processing: Volume 1-Volume 1}.\hskip 1em plus
  0.5em minus 0.4em\relax Association for Computational Linguistics, 2009, pp.
  296--305.

\bibitem{miller1956magical}
G.~A. Miller, ``The magical number seven, plus or minus two: Some limits on our
  capacity for processing information.'' \emph{Psychological review}, vol.~63,
  no.~2, p.~81, 1956.

\bibitem{aflalo2013spectral}
Y.~Aflalo and R.~Kimmel, ``Spectral multidimensional scaling,''
  \emph{Proceedings of the National Academy of Sciences}, vol. 110, no.~45, pp.
  18\,052--18\,057, 2013.

\end{thebibliography}

\end{document}